\documentclass[a4paper,fleqn]{cas-dc}

\usepackage[numbers]{natbib}
\usepackage{adjustbox}

\def\tsc#1{\csdef{#1}{\textsc{\lowercase{#1}}\xspace}}
\tsc{WGM}
\tsc{QE}
\tsc{EP}
\tsc{PMS}
\tsc{BEC}
\tsc{DE}

\begin{document}
\let\WriteBookmarks\relax
\def\floatpagepagefraction{1}
\def\textpagefraction{.001}
\shorttitle{Supervised Post-Processing for Person Detection}

\title [mode = title]{Choosing Supervised Post-Processing for Frozen Person Detectors: Information, Object Changes, and Complete Cost}                      



\author[1]{Tianyi Zhou}
\ead{tyzhou23-c@my.cityu.edu.hk}
\credit{Conceptualization, Methodology, Software, Investigation,
Validation, Writing -- original draft}

\affiliation[1]{organization={Department of Data Science, City University of Hong Kong},
                addressline={83 Tat Chee Avenue}, 
                city={Kowloon Tong},
                state={Hong Kong SAR},
                country={China}}

\author[2]{Zijun Jia}[orcid=0009-0005-7192-2592]
\cormark[1]
\ead{21375166@buaa.edu.cn}
\credit{Conceptualization, Methodology, Software, Formal analysis, Visualization, Writing -- review \& editing}



\affiliation[2]{organization={School of Automation Science and Electrical Engineering, Beihang University},
            addressline={No. 37 Xueyuan Road, Haidian District}, 
            state={Beijing},
            country={China}}



\cortext[cor1]{Corresponding author}


\setlength{\emergencystretch}{2em}
\renewcommand{\topfraction}{0.90}
\renewcommand{\textfraction}{0.10}
\renewcommand{\floatpagefraction}{0.80}
\newcommand{\IoU}{\operatorname{IoU}}
\newcommand{\UP}{\operatorname{UP}}

\begin{abstract}
A frozen detector computes hypotheses that can support better final
outputs through supervised post-processing. Choosing a readout requires
knowing which available information improves detection and what its
processing costs. We compare candidate scores, class distributions,
geometry, native-set summaries and adapted neighborhood meta classification
under common candidate, supervision and operating-point conditions. On
320 held-out CrowdHuman images, adding native-set context to a
97-dimensional readout yields 275 more matches and 96 more unmatched
predictions. On a separate 319-image cohort, Full100 matches 6004 objects,
versus 5848 for source-selected raw WBF and 6002 for a MetaDetect-style
adaptation. Its gain over raw contains 300 added and 144 lost discoveries.
After execution-equivalent optimization, Full100 and the meta-statistics
readout both take approximately 95 ms per complete warm request on a
fixed 64-image workload. Source-frozen transfer gives positive point
results on 223 MOT17Det frames from seven sequences, while the main
MIAP result shows no improvement. The study supports using sufficiently
informative output readouts while retaining optimized meta classification
as a competitive option. It also shows why feature dimension and net
detection gain alone are unreliable guides to adoption.
\end{abstract}


\maketitle

\section{Introduction}
A detector returns a small set of boxes after computing many more
hypotheses~\cite{tan2019learning}. Some discarded boxes localize real objects, whereas others
duplicate a detection or describe background~\cite{liu2019adaptive,chu2020detection}. Supervised post-processing
can reuse this evidence to improve the final output~\cite{tychsen2018improving}. Its practical value
depends on more than additional matches: different scoring and suppression
rules can lose previous discoveries, add unmatched output and require
extra processing after the image forward pass~\cite{bolya2020tide}.

We study this adaptation option with the original detector parameters
fixed. The interface exposes a finite pre-NMS person-candidate bank,
class probabilities and native detections from the same forward pass;
source-task labels are available for fitting and operating-point selection.
This defines the resources being compared. It does not imply that
fine-tuning is prohibited or inferior, and a final-box-only API or
label-free task would provide different resources. The question is how
much useful adaptation can be obtained from the detector's computed
outputs under this specified supervision and execution interface.

The question builds on established methods. MetaDetect predicts
detection quality from class scores, geometry and neighboring
hypotheses~\cite{schubert2021metadetect}. Detection MetaFusion uses
meta-classification confidence in post-processing~\cite{riedlinger2023gradient}.
Localization quality, learned suppression and compressed crowd outputs
provide further sources of selection evidence~\cite{jiang2018iou,hosang2017nms,wu2023crowdhat}.
These works establish the utility of quality prediction and output reuse.
Their different candidate, supervision and output protocols do not,
however, directly settle a choice at a common source-supervised interface:
which information changes the final policy, which strong alternative
provides the practical benefit, and what is paid for each execution path?

We compare these choices using the same person-candidate bank and matched
fitting and operating-point opportunities. Full100 combines 91 class
log-probability slots, six geometry terms and three summaries of the native
detection set. Input subsets and a five-dimensional score-plus-context
control test simplification. Adapted MetaDetect-style neighborhood
statistics test whether established meta classification explains the
quality. Raw processing receives its own source-selected threshold and
operator, while ROI and graph controls retain their additional resources.
Every complete policy is evaluated on absolute quality and changed
matched objects; complete time includes the information it actually uses.

The resulting choices are more specific than an improvement over a
detector's default output. Native-set context has a substantial conditional
contribution to the matched readout, whereas adding class channels alone
does not improve quality monotonically. The five-dimensional control
does not reproduce Full100's quality. Nevertheless, a strong
MetaDetect-style adaptation matches almost the same number of objects
and has slightly higher policy AP50 on the principal source cohort.
After execution-equivalent optimization, its complete time is also close
to Full100's. Input dimension is therefore an inadequate proxy for the
cost of obtaining and using that information.

Changes to the selected set also matter. On 319 CrowdHuman images,
Full100's 156 additional matches over optimized raw WBF comprise 300
gains and 144 losses. Two readouts with nearly identical totals can
replace many individual discoveries. These losses describe a policy's
behavior without imposing a requirement to preserve every previous box.
Source-frozen results further differ between MOT17Det and MIAP, so a
source-task gain alone does not determine transfer.

The study makes three contributions. First, it quantifies the conditional
value of candidate-local and native-set information under matched
supervision, learning configuration and operating-point opportunities.
Second, it compares the resulting adaptations with strong raw processing
and optimized neighborhood meta classification, including quality and
complete cost on the same images. Third, it characterizes the discoveries
and output burden exchanged by these policies across source cohorts,
a separately fitted detector and source-frozen evaluations. Together these
experiments support concrete choices about what information to acquire
and which post-processing alternatives to retain.

\begin{table*}[tbp]\centering\footnotesize
\caption{Direct precedents and the interfaces compared here. The executed
MetaStats policies adapt published feature families; the other cited
systems are methodological precedents rather than numerical reproductions.}
\label{tab:related}
\begin{tabular}{p{.19\linewidth}p{.32\linewidth}p{.39\linewidth}}\toprule
Work & Quality information / action & Relation to this study \\ \midrule
MetaDetect~\cite{schubert2021metadetect} & Class, geometry and neighboring hypotheses; quality estimation & Common-bank validity targets, learner and OP in MetaStats138/141 \\
Detection MetaFusion~\cite{riedlinger2023gradient} & Meta confidence in output filtering & Post-raw-NMS score replacement compared with learned ordering \\
IoU-Net~\cite{jiang2018iou} & Visual localization confidence; selection and refinement & Different visual/coordinate resources; no original-system run \\
Learning NMS~\cite{hosang2017nms} & Box relations and matching-dependent suppression & Graph/quality task control has additional supervision and dependencies \\
Crowd Hat~\cite{wu2023crowdhat} & Compressed output context; crowd-aware selection & Output reuse precedent; different crowd/counting targets \\
\bottomrule\end{tabular}
\end{table*}

\section{Related Work}
\subsection{Prediction quality from detector outputs}
MetaDetect constructs transparent metrics from the detector's class
probabilities, geometry and overlapping hypotheses, learning both
true/false-positive meta classification and IoU
regression~\cite{schubert2021metadetect}. It is the closest methodological
precedent for our output-statistics references. Riedlinger et al. study
gradient-based uncertainty and use detection meta-classification scores in
the MetaFusion application~\cite{riedlinger2023gradient}. That application
includes low-score candidate admission, raw-score suppression and
meta-score replacement. Our common-bank experiments distinguish such
replacement from using a learned ordering before suppression.

The evaluated MetaStats138 and MetaStats141 are transparent task
adaptations. They share the person bank, validity labels, learner and
operating-point procedure with Full100; they do not reproduce the original
authors' entire systems. In particular, adapting meta classification from
retained detections to a finite pre-NMS bank changes which hypotheses
can enter the final set. The shared interface makes the empirical
comparison interpretable without treating published numbers from other
detectors and metrics as comparable observations.

\subsection{Localization, competition and crowd evidence}
IoU-Net estimates localization confidence from visual representations for
selection and refinement~\cite{jiang2018iou}. Learning NMS learns
relations between boxes and scores with matching-dependent
targets~\cite{hosang2017nms}. Crowd Hat compresses detector-output area
and confidence information to estimate crowd context and adapt selection,
with fixed detector weights~\cite{wu2023crowdhat}. These works establish
that computed hypotheses and their relations can be useful beyond the
original confidence score.

IoU-aware calibration also conditions confidence on competition with
higher-scoring same-class boxes and models duplication~\cite{gilg2024iou}.
Its duplicate-aware objective differs from our independent candidate
validity labels, for which several boxes around one person can be
positive. Hard-NMS subsequently resolves their competition. Soft-NMS
changes scores during suppression~\cite{bodla2017soft}, while weighted
boxes fusion (WBF) can change representative coordinates as well as
scores~\cite{solovyev2021wbf}. These are substantive alternatives, rather
than interchangeable implementations of the same output rule.

\subsection{Calibration and complete-policy evaluation}
Detection calibration can include location and scale in addition to
confidence~\cite{kuppers2020multivariate}. Its evaluation depends on
matching, threshold and processing choices~\cite{kuzucu2024calibration,huseljic2025systematic}.
We use scalar logistic calibration as part of a source operating-point
procedure, and evaluate the complete thresholded policy. The calibration
and bootstrap procedures are standard tools. Our focus is their consistent
use across representations, including changed matched-object sets and
the processing needed to obtain each representation.

\begin{figure*}[tbp]\centering
\includegraphics[width=\linewidth]{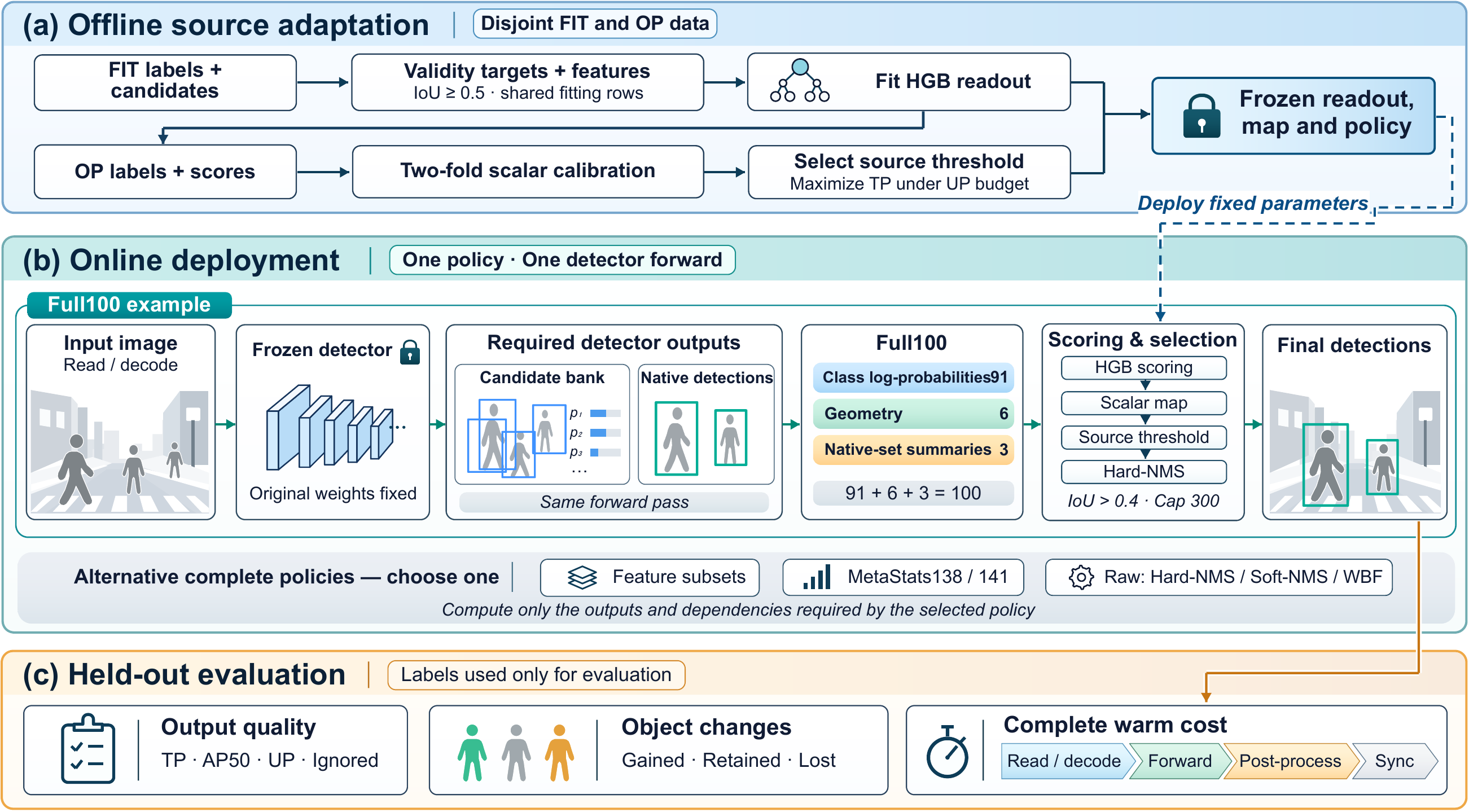}
\caption{Study and deployment interface. (A) Source labels train readouts
and select scalar maps and thresholds offline. (B) A request executes one
chosen policy with one fixed-detector forward and its required outputs.
Full100 and meta-statistics readouts are alternatives. (C) Held-out
evaluation compares complete outputs and processing costs.}\label{fig:workflow}
\end{figure*}

\section{Candidate-to-Output Adaptation}
\subsection{Interface and candidate supervision}
For image $I_i$ of width $W_i$ and height $H_i$, the fixed detector supplies
\begin{equation}
C_i=\{(b_{ij},p_{ij},s_{ij})\}_{j=1}^{n_i},\qquad D_i=\{(d_{ik},c_{ik})\}_k.
\end{equation}
Here $b_{ij}$ is a normalized person-regressed candidate box, $p_{ij}$
is its class-probability vector and $s_{ij}$ is its original person
score. $D_i$ is the native person output. The bank admits valid candidates
at original score at least 0.005 and retains at most 1000 in raw-score
order. The feature extractor, classifier and box regressor retain their
original parameters. A service exposing only final boxes would need a
different interface.

Let $G_i$ contain eligible person annotations and $U_i$ ignored regions.
The independent candidate-validity target is
\begin{equation}
y_{ij}=\mathbf{1}\{\max_{g\in G_i}\IoU(b_{ij},g)\geq0.5\},
\label{eq:label}
\end{equation}
with maximum zero for an empty eligible set. A remaining negative is
excluded from fitting if its intersection-over-candidate-area with an
ignore region is at least 0.5. Eligible positives take precedence over
ignore. The target intentionally permits duplicate positives: it estimates
whether a candidate can localize a person, rather than assigning exclusive
ownership of an object. The final output rule must still resolve duplicates.

\subsection{Information groups and Full100}
The feature vector is
\begin{equation}
x_{ij}=[\ell_{ij};\gamma_{ij};\nu_{ij}]\in\mathbb{R}^{100}.
\end{equation}
$\ell_{ij}$ contains 91 log probabilities, including background and unused
category slots, computed as $\log\operatorname{clip}(p,10^{-7},1)$.
$\gamma_{ij}$ contains four normalized coordinates, log normalized area
and log pixel aspect ratio. The final three entries summarize agreement
with the native set. With $o_k=\IoU(b_{ij},d_{ik})$ and $k^*$ the first
argmax in native output order,
\begin{equation}
\nu_{ij}=\left[\max_k o_k,\ c_{ik^*},\
\max_{k:o_k\geq0.1} c_{ik}\right].\label{eq:native}
\end{equation}
An empty native set gives three zeros, and an empty qualifying set in
the last entry gives zero. If a nonempty native set has all-zero overlap,
the second entry still uses its first confidence. This edge convention is
retained in all matched controls.

Score2 uses only background/person log probabilities; Class91 uses all
91 slots; ClassGeom97 adds geometry; Full100 adds the native summaries.
Score2Native3, abbreviated Score5 in figures, uses the two log probabilities
and three summaries. These feature groups describe fitted policies rather
than a sequence of guaranteed improvements. The full table of columns and
numerical floors appears in Appendix S1.

Native-set information is computed in the same detector forward path.
Native output first applies multiclass NMS and a total cap of 300, then
filters person detections at score 0.4. Adapted output can retain 300
person boxes. Raw-bank controls account for this difference from the native
interface. Full100 exports no additional ROI embedding, although Faster
R-CNN performs its usual ROI computation internally.

\subsection{Fitting, calibration and operating points}
A histogram gradient-boosted classifier (HGB)~\cite{friedman2001gbm,pedregosa2011sklearn}
fits the validity labels using log loss. It has 120 iterations, learning
rate 0.1, at most 15 leaves, $L_2=2$, and no early stopping. Each seed
uses all eligible positive rows and at most 150,000 sampled negatives.
The main sample contains 254,141 positives and 150,000 negatives. Matched
input groups share row identities, order, labels and HGB settings; the
available predictors, fitted trees and effective capacity still differ.
Seed 25 defines the principal instance and paired seeds 26/27 assess
training variability.

Source fitting (FIT), operating-point selection (OP) and adaptive
development (DEV) use distinct images. Two hash-defined OP folds generate
out-of-fold scores: a scalar logistic map fitted on one fold scores the
other. For predicted validity $r$ the map is
\begin{equation}
\widetilde r=\sigma\big(a\operatorname{logit}
(\operatorname{clip}(r,10^{-6},1-10^{-6}))+b\big).
\end{equation}
A map fitted on the full OP set is deployed. Each policy selects its
own threshold to maximize OP TP under a common UP ceiling, 2388 at the
principal budget, equal to native OP UP. Atomic equal-score events can
leave budget unused. Equal OP opportunity does not imply equal realized
UP on confirmation data. Appendix S2 specifies ties, scalar fitting
and output rules.

\subsection{Inference and strong alternatives}
At inference, a single detector forward supplies the required candidate
and native outputs. CPU feature construction, HGB prediction and the
full-OP scalar map yield the new scores. Boxes are clipped to $[0,1]$,
invalid/non-finite rows and scores below threshold are removed, and stable
score ordering precedes greedy Hard-NMS. Main learned policies suppress
at IoU $>0.4$ and retain at most 300 boxes. Candidate order breaks score
ties. No GT, second detector or additional image encoding is required.
Figure~\ref{fig:workflow} separates source adaptation from this online path.

MetaStats138 adapts MetaDetect's statistic families to the same person
bank: 98 basic score/class/pixel-geometry entries and 40 neighborhood
statistics. A neighborhood includes the anchor and candidates with IoU
$>0.45$. MetaStats141 appends the same native summaries. Both use matched
labels, sampled rows, HGB configuration, OP and Hard-NMS. This common-bank
adaptation changes targets and numerical geometry from the original
system; Appendix S1 details those differences. A post-NMS variant
instead suppresses with raw scores at IoU 0.5, replaces survivor scores
with meta scores and applies its own threshold. It is a complete-policy
alternative, not an isolated causal test of suppression order.

Raw Hard-NMS, Gaussian Soft-NMS and WBF receive source-OP selection
without learned quality scores. A fixed raw Hard-0.4 policy additionally
matches the principal output operator. ROI132 and a graph/quality control
test stronger same-forward representations, with all projection and
scorer dependencies charged. An independently fitted RetinaNet readout
tests a different detector interface; it is not transfer of the two-stage
head. Their additional information and output dependencies are specified
alongside their quality results.

\begin{table}[!t]\centering\small
\renewcommand{\arraystretch}{0.94}
\caption{Quality of the original frozen policies on the full confirmation cohorts. CH-319 has 7246 eligible visible-body annotations; MOT17Det has 4491 eligible full-body annotations in 223 frames from seven sequences. AP50 uses the CH author-visible and custom MOT native-box protocols separately. $G/L$ are objects gained/lost by Full100 relative to each row. The label-free implementation comparison in Section~\ref{sec:cost} concerns a separate 64-image subset.}\label{tab:newconfirmation}
\adjustbox{width=1\linewidth}{
\begin{tabular}{lrrrrr}\toprule
Policy & TP & UP & AP50 (\%) & $G$ & $L$ \\ \midrule
\multicolumn{6}{l}{CrowdHuman, 319 images} \\
Native & 5684 & 1342 & 74.19 & 430 & 110 \\
Raw Hard .4 & 5802 & 1348 & 75.92 & 316 & 114 \\
Raw OP-best WBF & 5848 & 1387 & 76.77 & 300 & 144 \\
Full100 & 6004 & 1366 & 80.10 & 0 & 0 \\
Score2Native3 & 5804 & 1385 & 76.05 & 290 & 90 \\
MetaStats138 & 6002 & 1392 & 80.16 & 154 & 152 \\
MetaStats141 & 5990 & 1387 & 80.00 & 147 & 133 \\
Post-NMS 138 & 5719 & 1406 & 76.22 & 415 & 130 \\
Post-NMS 141 & 5856 & 1389 & 78.27 & 280 & 132 \\
\midrule
\multicolumn{6}{l}{MOT17Det, 223 frames / seven sequences} \\
Native & 2981 & 1350 & 61.26 & 121 & 72 \\
Raw Hard .4 & 2994 & 1492 & 61.45 & 106 & 70 \\
Raw OP-best WBF & 2988 & 1570 & 61.34 & 127 & 85 \\
Full100 & 3030 & 1349 & 63.71 & 0 & 0 \\
Score2Native3 & 2973 & 1456 & 60.80 & 115 & 58 \\
MetaStats138 & 3000 & 1334 & 62.71 & 94 & 64 \\
MetaStats141 & 3010 & 1348 & 63.18 & 83 & 63 \\
Post-NMS 138 & 2910 & 1128 & 61.08 & 191 & 71 \\
Post-NMS 141 & 2943 & 1163 & 61.80 & 164 & 77 \\
\bottomrule\end{tabular}
}
\end{table}

\section{Experimental Protocol}
\subsection{Detectors, annotations and data roles}
The primary network is TorchVision Faster R-CNN R50-FPNv2 with fixed
COCO\_V1 weights~\cite{ren2015faster,lin2017fpn,torchvision_faster_v2},
short side 1200 and maximum side 2000. The dense control uses RetinaNet
R50-FPNv2~\cite{lin2017focal,torchvision_retina_v2} with separate source
fitting. CrowdHuman visible-person annotations define the source
task~\cite{shao2018crowdhuman}. Source FIT/OP/DEV contain 1024/512/256
records. Person targets and ignore semantics are adapted explicitly;
in particular, candidate positives use visible boxes and training ignore
regions use full-body boxes.

The evaluations address distinct questions. CH-255 tests representation
controls, Info-320 tests the four information groups, and CH-319 tests
the direct output-statistics references. Their 5332, 7278 and 7246
eligible objects belong to separate same-source cohorts, not three
external datasets. MIAP-256 uses 1119 eligible native-person boxes from
MIAP/Open Images~\cite{schumann2021miap,kuznetsova2020openimages}.
MOT17Det contributes 223 frames and 4491 eligible full-body objects
from seven videos~\cite{mot17det}. Source-fitted heads and thresholds
remain unchanged on both external sources.

\begin{table*}[!t]\centering\small
\caption{Full100 minus each primary reference in the new cohorts. Individual 98.75\% intervals target approximate 95\% family coverage separately within each source. CH resamples paired images; MOT resamples paired contributions in seven sequence blocks. Negative UP differences indicate fewer unmatched predictions.}\label{tab:newintervals}
\begin{tabular}{llrr}\toprule
Source / reference & Endpoint & Difference & Interval \\ \midrule
CH / Score2Native3 & Recall (pp) & +2.760 & [+2.026, +3.496] \\
CH / Score2Native3 & UP/image & -0.060 & [-0.323, +0.207] \\
CH / MetaStats141 & Recall (pp) & +0.193 & [-0.379, +0.769] \\
CH / MetaStats141 & UP/image & -0.066 & [-0.266, +0.129] \\
MOT17 / Score2Native3 & Recall (pp) & +1.269 & [+0.507, +2.447] \\
MOT17 / Score2Native3 & UP/image & -0.480 & [-1.000, +0.080] \\
MOT17 / MetaStats141 & Recall (pp) & +0.445 & [-0.377, +0.949] \\
MOT17 / MetaStats141 & UP/image & +0.004 & [-0.420, +0.371] \\
\bottomrule\end{tabular}
\end{table*}

Record identities, available image hashes and metadata were screened
against previous roles before the respective confirmation labels were
exposed. One planned record was excluded without replacement in CH-255,
CH-319 and MOT17Det. All twelve direct-reference policies and primary
contrasts were fixed before the CH-319/MOT annotations were evaluated;
both cohorts' predictions were completed first. Record disjointness does
not prove independent scenes or eliminate unknown pretraining overlap.
The cohort table and screening details are in Appendix S3.

\subsection{Quality and individual discoveries}
The count protocol matches eligible annotations one-to-one in descending
prediction-score order at IoU $\geq0.5$. An unmatched prediction is ignored
at intersection-over-prediction-area $\geq0.5$ with an ignore region;
otherwise it contributes to UP. Thus
\begin{equation}
\mathrm{outputs}=\mathrm{TP}+\mathrm{UP}+\mathrm{ignored}.
\end{equation}
UP is an annotation-based burden measure, not a human-adjudicated count of
physical false positives. Ignored outputs remain part of actual output.

For a policy $a$ and reference $r$, let $T_i(a)$ and $T_i(r)$ be their
matched eligible object identities. We report
\begin{equation}
G_i(a;r)=|T_i(a)\setminus T_i(r)|,\quad
L_i(a;r)=|T_i(r)\setminus T_i(a)|.\label{eq:changes}
\end{equation}
Then $G-L$ equals the TP difference. This descriptive decomposition
prevents new matches from concealing regressions of other objects. It
depends on the specified matcher, rather than establishing a unique
physical assignment in ambiguous overlaps.

Policy AP50 evaluates the thresholded and capped output. CrowdHuman uses
author-code visible-box AP with strict IoU $>0.5$ and consecutive
trapezoid integration; its ignore geometry differs from the count protocol.
MIAP and MOT use native-box precision-envelope integration, with the
latter a custom adaptation rather than the official MOT score. Results
from these protocols are reported separately, not averaged as a common AP.

\subsection{Uncertainty and complete processing cost}
The CH-319 and MOT studies each prespecify four endpoints: Full100 minus
Score5 and MetaStats141, each on pooled recall and UP/image. CH uses
30000 paired-image bootstrap draws; MOT uses 30000 draws in seven
sequence blocks. Individual 98.75\% intervals target approximate 95\%
Bonferroni family coverage separately within each source. With only seven
MOT blocks, interval interpretation is necessarily limited. The earlier
cohorts retain their own families. AP and object changes are secondary
outcomes. An interval crossing zero does not establish equivalence or
non-inferiority.

\begin{table}[!t]
\centering\small
\caption{Info-320 results on 7278 eligible visible-body annotations. The four input groups use their prespecified seed-25, budget-$B$, Hard-NMS .4 policies. Native and raw anchors are unchanged. AP50 uses the author-visible protocol.}\label{tab:infoconfirm}
\begin{tabular}{lrrr}
\toprule
Policy & TP & UP & Policy AP50 (\%) \\
\midrule
Native & 5708 & 1464 & 74.23 \\
Raw Hard .4 & 5795 & 1463 & 75.53 \\
Raw OP-best WBF & 5856 & 1443 & 76.53 \\
Score2 & 5767 & 1423 & 75.21 \\
Class91 & 5708 & 1424 & 75.27 \\
ClassGeom97 & 5719 & 1408 & 75.48 \\
Full100 & 5994 & 1504 & 79.41 \\
\bottomrule
\end{tabular}
\end{table}

Complete warm-request time starts at image read and decoding and ends
after synchronized final output. It includes candidate capture, required
transfers, feature construction, all heads and post-processing. Every
policy performs its own real forward, skipping unused dependencies.
Three interleaved repetitions are averaged within each image. The final
execution study uses the same 64 images, 32 CH and 32 MOT, in two
separately initialized jobs with reversed policy order. Its quality and
time refer to those same outputs. Warmup, loading, source annotation,
fitting and data preparation are separate resources. Appendix S4 gives
the implementation-equivalence checks, hardware and earlier sessions.

\section{Results}
\subsection{What remains after strong raw and meta-classification controls?}
Table~\ref{tab:newconfirmation} reports frozen policies on CH-319 and
MOT17Det. On CH-319, Full100 matches 6004 objects, compared with 5848
for source-selected raw WBF and 5802 for raw Hard-0.4. Its 1366 UP
lie between these raw policies' counts. The gain is therefore more than
releasing low-score candidates or retaining the native settings.
MetaStats138 reaches 6002 TP and a slightly higher policy AP50
(80.16\% versus 80.10\%), while MetaStats141 reaches 5990 TP and
1387 UP. These points support supervised adaptation with multiple
competitive readouts.

\begin{figure*}[tbp]\centering
\includegraphics[width=0.7\linewidth]{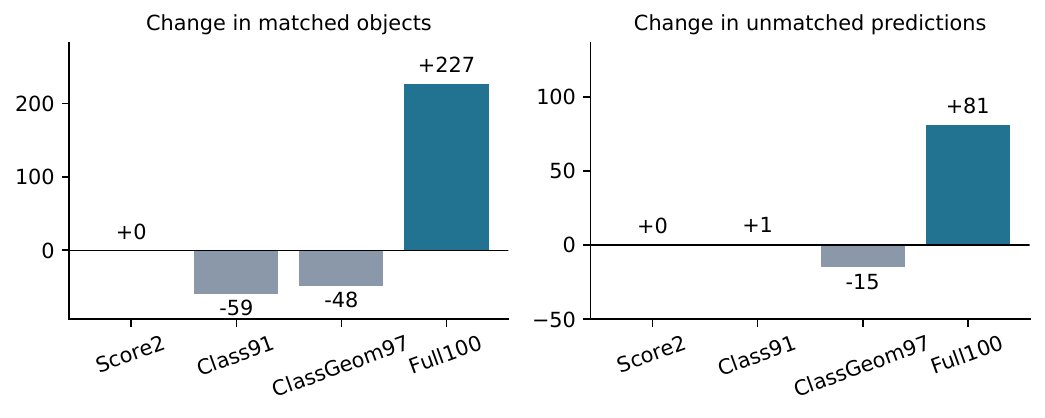}
\caption{Matched information groups on Info-320. TP and UP are shown
relative to Score2. Fitting rows and Hard-NMS are shared, while each
policy has its own source-selected threshold.}\label{fig:information}
\end{figure*}

\begin{figure*}[tbp]\centering
\includegraphics[width=0.7\linewidth]{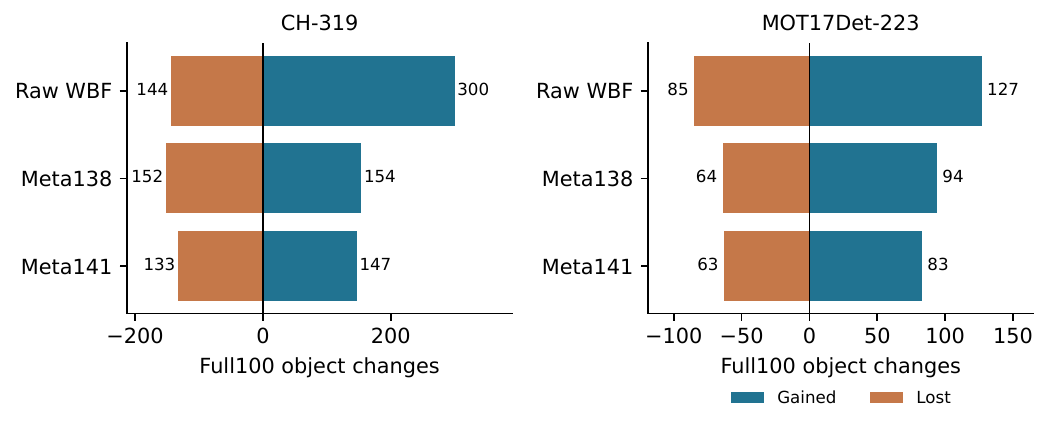}
\caption{Objects gained and lost by Full100 relative to strong references.
Each row uses actual matched annotation identities; CH-319 and MOT17Det
are distinct cohorts.}\label{fig:changes}
\end{figure*}

\begin{figure*}[tbp]\centering
\includegraphics[width=0.6\linewidth]{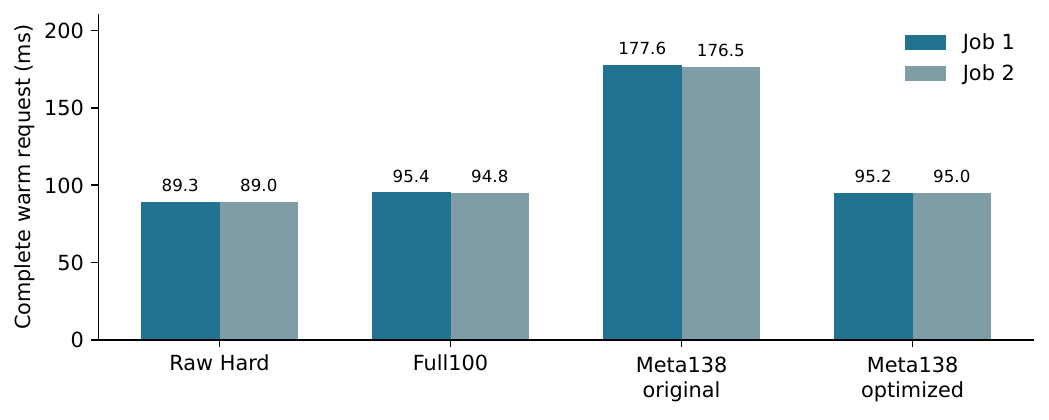}
\caption{Complete warm-request means on the fixed 64 images. Original
and execution-equivalent optimized Meta138 are compared with Full100
and raw processing. Jobs are repeated engineering measurements.}\label{fig:cost}
\end{figure*}

\begin{table}[!t]\centering\footnotesize
\caption{Quality and complete warm-request time on the same 64 images
(32 CH and 32 MOT; 1253 eligible objects). Each job averages three
requests per image. Quality is identical across the two jobs and is
linked to their saved output identities. Times are descriptive means
in milliseconds; no subset AP or new performance test is computed.}
\label{tab:same64}
\setlength{\tabcolsep}{4pt}
\adjustbox{width=1\linewidth}{
\begin{tabular}{lrrrrrr}\toprule
Policy & TP & UP & Ignored & Outputs & Job 1 & Job 2 \\ \midrule
Raw Hard & 897 & 348 & 332 & 1577 & 89.276 & 89.044 \\
Score5 & 899 & 338 & 345 & 1582 & 94.220 & 92.947 \\
Full100 & 925 & 308 & 369 & 1602 & 95.374 & 94.765 \\
Meta138 optimized & 922 & 311 & 369 & 1602 & 95.210 & 95.038 \\
Meta141 optimized & 924 & 309 & 360 & 1593 & 96.776 & 95.529 \\
\bottomrule\end{tabular}
}
\end{table}

\begin{table}[!t]\centering\small
\caption{Separately fitted RetinaNet readout on CH-255 (5332 objects).
Raw Gaussian is the source-selected raw alternative; raw WBF matches
the learned output operator. AP50 uses author-visible boxes.}
\label{tab:retinaprimary}
\begin{tabular}{lrrr}\toprule
Policy & TP & UP & Policy AP50 (\%) \\ \midrule
Raw Gaussian & 3520 & 507 & 63.68 \\
Raw WBF & 3489 & 486 & 63.29 \\
Validity WBF & 3691 & 566 & 67.47 \\
\bottomrule\end{tabular}
\end{table}

\begin{table*}[!t]\centering\footnotesize
\caption{Adoption choices supported by the evaluated interface and
workloads. These recommendations concern the observed policies, not
rules predicting success on an arbitrary new source.}\label{tab:adoption}
\begin{tabular}{p{.22\linewidth}p{.43\linewidth}p{.25\linewidth}}\toprule
Choice & Evidence-backed implication & Scope \\ \midrule
Before learning & Give released-bank raw scores a source OP and operator comparison & Source labels and candidate access required \\
Simplify the input? & Score5 leaves a quality gap with only a small complete-time saving on the fixed64 & No minimal-sufficiency claim \\
Full100 or Meta? & Retain both; optimized Meta138 is close in quality and time, with different dependencies & No equivalence or universal winner \\
Add appearance? & ROI/graph points improve some quality coordinates; charge all added dependencies & Stronger-resource alternatives \\
Change source? & Check intended-source behavior; MOT points and MIAP failure differ & No general transfer rule \\
\bottomrule\end{tabular}
\end{table*}

The five-dimensional score-plus-context control reaches 5804 TP and
1385 UP. Full100's 200-match advantage has a positive prespecified
recall interval, while the UP interval includes zero
(Table~\ref{tab:newintervals}). Against MetaStats141 the recall interval
also includes zero. The five inputs are insufficient to reproduce the
rich readout's quality here, but the neighborhood-statistics adaptation
accounts for most of that quality. Neither superiority nor equivalence
to MetaStats141 is established by the interval.

On DEV-256, Full100 reaches 4591 TP/1141 UP versus MetaStats138's
4578/1120 and MetaStats141's 4585/1107. Post-NMS replacement reaches
4341/1111 and 4445/1077. Learned pre-NMS ordering is favorable at
these selected points, but the alternatives also differ in raw-NMS
setting and threshold selection. Appendix S3 retains all seeds and
budgets as complete-policy development comparisons.

\subsection{Which available information changes the policy?}
Info-320 holds candidate identities, target labels, fitting rows, HGB
configuration, OP opportunity and Hard-NMS fixed. Quality is not monotone
in feature count (Table~\ref{tab:infoconfirm}). Class91 matches fewer
objects than Score2; geometry partly closes that gap. Adding native-set
summaries gives the largest conditional increment. Full100 exceeds Score2
by 227 TP and ClassGeom97 by 275 TP, while the latter comparison
also adds 96 UP and 735 ignored outputs.

The prespecified recall differences are +3.119 percentage points against
Score2 (98.75\% interval [2.175,4.053]) and +3.779 against
ClassGeom97 ([2.739,4.857]). UP/image changes by +0.253
([-0.016,0.531]) and +0.300 ([0.006,0.597]). The gain therefore
has a burden tradeoff. Its TP direction is consistent across the three
paired seeds and three OP budgets on DEV, which are related sensitivity
points rather than independent tests.

These results favor native-set context in this readout, without separating
the causal effects of its three entries. Inputs, fitted trees, scalar maps
and selected thresholds change together; confirmation UP is not forced
equal. Figure~\ref{fig:information} displays the resulting policy effects.
Together with the five-dimensional and MetaStats controls, the comparison
supports sufficiently rich output evidence rather than a uniquely
privileged feature dimension.

\subsection{Net improvement and changed discoveries}
On CH-319, Full100 gains 300 objects and loses 144 relative to raw WBF.
Against MetaStats138 it gains 154 and loses 152, despite a net difference
of only two matches; against MetaStats141 it gains 147 and loses 133.
On Info-320, its 275 additional TP over ClassGeom97 contain 382 gains
and 107 losses. Figure~\ref{fig:changes} shows the magnitude of this
replacement. Greater TP does not imply a superset of previous discoveries.

UP alone also misses changes in ignored output. On CH-319, Full100 emits
334 more predictions than raw WBF: 156 additional TP, 21 fewer UP and
199 additional ignored predictions. On Info-320, Full100 emits 9440
predictions, including 1942 ignored outputs, while ClassGeom97 emits
8334, including 1207 ignored outputs. The extra 1106 predictions contain
different useful, unmatched and ignored components. Appendix S3 gives
the complete accounting without assigning a universal cost to each category.

\subsection{Quality and complete cost on the same images}\label{sec:cost}
We compare readouts after optimizing the neighborhood-statistics execution.
The C99 path preserves the frozen features, scores, threshold membership
and ordered outputs on the checked inputs. Synthetic and real-packet
comparisons establish this execution equivalence for the measurement
workload; the quality models and policies remain unchanged.

On the fixed 64 images, Full100 takes 95.374 and 94.765 ms in two
separately initialized jobs, versus 95.210 and 95.038 ms for optimized
MetaStats138. The paired mean differences are +0.163 and $-0.273$ ms,
reversing direction. These workload observations remove the large cost
separation seen with the original implementation; they are not an
equivalence test. Optimized MetaStats141 takes 96.776 and 95.529 ms,
leaving Full100 a smaller mean advantage of 1.403 and 0.764 ms.

Table~\ref{tab:same64} makes the practical comparison explicit. Score5
is only 1.154 and 1.819 ms faster than Full100, but matches 26 fewer
objects and has 30 more UP on those same images. Full100 and Meta138
are much closer in both quality and complete time. A small input vector
does not automatically provide a favorable economy after capture,
native processing, transfer and scoring are included.

The supervised overhead relative to raw remains real: Full100 adds
6.097 and 5.721 ms, or 6.83\% and 6.42\% by ratios of mean times.
Every policy in Table~\ref{tab:same64} has its own forward and the quality
of exactly those outputs. No timing-subset AP is computed, and full-cohort
CH-319/MOT quality is not paired with this subset clock. Loading,
annotation, fitting, cold starts and sustained throughput lie outside
the warm-request measurement.

\subsection{Source and detector changes}
The earlier CH-255 source confirmation also favors supervision: Full100
reaches 4461 TP/1071 UP versus raw WBF's 4371/1057, with 193 gains
and 103 losses. ROI132 reaches 4472/1065 and graph/quality reaches
4482/1072 with their additional representation and scorer dependencies
(Appendix S1 and S3).

MOT17Det gives a positive source-frozen point: Full100 reaches
3030 TP/1349 UP versus raw Hard's 2994/1492. Its 20 net matches
over MetaStats141 comprise 83 gains and 63 losses. The sequence-level
TP differences are +12, +11, +1, +1, $-4$, +1 and $-2$; two sequences
account for more than the pooled net gain. The primary block interval
crosses zero, and the 223 frames represent only seven temporal clusters.

MIAP-256 does not establish improvement for the main source-fitted
instance: Full100 has 933 TP/845 UP, versus raw WBF's 934/839 and
Native's 942/787. The recall difference to raw WBF is $-0.089$
percentage points (98.75\% interval [$-1.903$,1.579]), establishing
neither benefit nor equivalence. Image distribution, annotation coverage
and box geometry all change, so this evaluation does not isolate a cause.

The separately fitted RetinaNet interface shows a second source-task
benefit (Table~\ref{tab:retinaprimary}). On CH-255, quality-WBF gains
202 TP and 80 UP over raw WBF, and 171 TP and 59 UP over the
stronger source-selected raw Gaussian policy. On MIAP it reaches
831/336 versus raw Gaussian's 817/361. These results belong to the
dense detector's own fitted sigmoid-output head and operating point,
not transfer of the two-stage Full100 head.

An existing semantic sensitivity also retains source improvement:
the fixed CH-255 Full100 predictions have full-body author AP50 of
63.63\%, versus 57.54\% for raw WBF. This secondary evaluation changes
the annotation target without refitting (Appendix S3). It does not
explain the MIAP outcome. RetinaNet's complete WBF path is expensive,
even with matched raw/learned operators; its clocks remain paired with
their own images in Appendix S4.

\section{Discussion}
\subsection{Implications for adopting candidate reuse}
Table~\ref{tab:adoption} translates the evidence into choices under the
studied interface. A strong raw operating point establishes what candidate
release and standard processing already provide. When source supervision
is available, rich readouts are then worth comparing. The information
experiments favor sufficiently informative output evidence here, while
the strong MetaStats results show that this evidence need not take one
specific form. Full100 and optimized MetaStats138 remain competitive
options with different dependencies.

Full100 requires native-set processing and full class-probability transfer;
MetaStats138 requires neighborhood statistics but skips native processing.
ROI and graph policies add other dependencies. All are charged in the
corresponding request path. If a service exposes or charges different
information, its complete path needs measurement. Input dimension alone
does not specify this resource boundary.

Object changes and output accounting further constrain adoption. Similar
TP totals do not preserve the same annotations, and a decrease in UP
need not reduce ignored or total output. These consequences of changing the selection policy can inform an application's
requirements without assuming that every old discovery must be retained
or assigning a synthetic price to each error.

\subsection{Scope and limitations}
The experiments concern person detection with finite candidate banks and
source supervision, using two COCO-pretrained detector interfaces. They
do not establish a general preference across categories, detector families
or fine-tuning strategies. Input comparisons change fitted functions and
thresholds with the representation; they estimate conditional policy
effects rather than independent causal contributions of channels.
Candidate-validity labels permit redundant positives, leaving final
competition to the selected output operator.

Source confirmations are record-disjoint within one dataset. MOT evidence
is clustered in seven sequences, while MIAP differs in image distribution
and annotation semantics. Metadata screening cannot exclude all scene or
pretraining overlap. The different transfer outcomes support checking the
intended source, not predicting which new domain will succeed. Target-domain
fitting would be a separate adaptation study.

UP and object identities inherit annotation and matching choices,
including ambiguities in crowded overlaps. The time measurements describe
fixed shared-hardware workloads, not future-service or whole-lifecycle
savings. The performance intervals provide uncertainty for their stated
comparisons rather than deployment-risk bounds~\cite{jia2026balancerag,hu2026mird,jia2025ser}. Within these limits,
the study identifies useful source-task adaptation and the strong
alternatives and resources needed to evaluate it.

\section{Conclusion}
Computed detector candidates support useful source-task adaptation with
the original network unchanged. Under matched fitting and selection,
native-output context adds a conditional benefit beyond local class and
geometry, while optimized neighborhood meta classification remains a
competitive quality--cost option. Execution-equivalent optimization largely
removes the apparent efficiency separation between these alternatives.
A practical comparison begins with strong raw processing and retains
informative readouts alongside optimized meta classification. Complete
cost, changed objects and output burden then inform the choice.
The MOT17Det and MIAP outcomes show why the intended source and annotation
setting remain part of that choice.

\bibliographystyle{cas-model2-names}

\bibliography{cas-refs}

\appendix

\renewcommand{\thesection}{S\arabic{section}}
\renewcommand{\thetable}{S\arabic{table}}
\renewcommand{\thefigure}{S\arabic{figure}}

\section{Feature Interfaces and Fitted Readouts}
\subsection{Primary candidate capture and Full100}
The primary TorchVision Faster R-CNN R50-FPNv2 COCO\_V1
network~\cite{ren2015faster,lin2017fpn,torchvision_faster_v2} uses short
side 1200 and maximum side 2000. Candidate capture retains the
person-regressed box for rows with original person score at least 0.005,
with valid-box filtering and stable raw-score top 1000. It does not
require the person slot to be the largest class probability. Native output
uses score floor 0.01, classwise NMS 0.5 and a total multiclass cap of
300, followed by person filtering at score 0.4. Consequently the adapted
300-person cap is distinct from the native multiclass cap.

The same-forward feature array is float32. Table~\ref{tab:features}
specifies all columns. Probabilities are clipped before logarithms, and
the background and unused category slots are retained. The aspect ratio
uses pixel width and height, whereas area is normalized. These quantities
are not additional ROI appearance embeddings.

Eligible FIT positives are valid, non-ignored CrowdHuman person visible
boxes. A candidate is positive at maximum eligible IoU $\geq0.5$; empty
eligible sets have maximum zero. Only remaining negatives are removed
when their intersection-over-candidate-area with an ignored full-body
region is $\geq0.5$. Positive precedence and duplicate positives are
important: the training target is not the exclusive final matching target.

Each seed uses all 254141 eligible positives and the same seed-specific
150000 negative rows, selected from 409582 usable FIT candidates.
Input-group comparisons share row order, identities and labels. HGB has
log loss, learning rate 0.1, 120 iterations, at most 15 leaves,
minimum 20 samples per leaf, $L_2=2$ and maximum 255 bins. There is
no depth limit, class weighting, monotonic constraint or interaction
constraint; all features are available at each split. Early stopping
and warm starts are disabled. Seed 25 is primary; seeds 26 and 27
are paired stability runs, not independent data replications.
The same previously fitted Full100 heads are used in all comparisons.

\begin{table*}[tbp]\centering\small
\caption{Full100 features (zero-based columns). Coordinates $x_1,y_1,x_2,y_2$ are in pixels; $b$ denotes their normalization by image width $W$ and height $H$. Width $w_n$ and height $h_n$ are floored at $10^{-9}$. $D_i$ is the native person set in the same normalized coordinates.}\label{tab:features}
\begin{tabular}{p{.17\linewidth}p{.73\linewidth}}\toprule
Columns & Definition \\
\midrule
0--90 & $\log\operatorname{clip}(p_k,10^{-7},1)$; background is slot 0, person slot 1. \\
91--94 & $(x_1/W,y_1/H,x_2/W,y_2/H)$. \\
95 & $\log\max(w_nh_n,10^{-12})$. \\
96 & $\log\max(w_nW/(h_nH),10^{-9})$. \\
97 & $\max_{d\in D_i}\IoU(b,d)$. \\
98 & Confidence of the first argmax-IoU box in native output order. \\
99 & Maximum native confidence among boxes with IoU $\geq0.1$, or zero if none qualifies. \\
\bottomrule\end{tabular}
\par\smallskip\raggedright The three context entries are zero if $D_i$ is empty. If $D_i$ is nonempty but all overlaps are zero, column 98 returns the first native confidence rather than zero.
\end{table*}

\begin{table*}[tbp]\centering\small
\caption{Readout inputs and required processing. Every row starts with the same person-candidate bank. The detector is fixed; learned readouts require source FIT/OP supervision. Native context is computed from native boxes and scores, rather than supplied free of cost. Exact feature definitions are in Supplement S1.}\label{tab:inputs}
\begin{tabular}{p{.22\linewidth}p{.34\linewidth}p{.33\linewidth}}\toprule
Readout & Readout information & Required dependencies \\
\midrule
Raw controls & Original person score & Selected raw operator \\
Score2Native3 & Two log probabilities, three native summaries & Two probability slots; native processing; scalar/head \\
Full100 & 91 log probabilities, six geometry terms, three native summaries & All probability slots; native processing; scalar/head \\
MetaStats138 & 98 basic outputs and 40 neighborhood statistics & All probability slots; candidate-neighborhood computation; scalar/head \\
MetaStats141 & MetaStats138 plus native summaries & MetaStats138 dependencies plus native processing \\
\bottomrule\end{tabular}
\end{table*}

\subsection{Adapted MetaDetect and MetaFusion references}
MetaStats138 adapts MetaDetect's transparent statistic
families~\cite{schubert2021metadetect}. Its 98 base entries comprise
four pixel coordinates, raw person score, constant person category 1,
sum of class probabilities, and 91 class probabilities. The latter
are reconstructed by exponentiating the same float32 log slots used
by Full100, after conversion to float64. For normalized boxes $b_i$,
the anchor-relative neighborhood is
\begin{equation}
N_i=\{j:\IoU(b_i,b_j)>0.45\}\cup\{i\}.
\end{equation}
The 40 further entries are: neighborhood size (1); minimum, maximum,
mean and population standard deviation of center coordinates and
half-width/half-height (16); anchor value and four neighborhood
statistics for area, perimeter and their ratio (15); four raw-score
statistics (4); and four positive-overlap statistics (4).

Pixel width/height have floor $10^{-9}$, perimeter is twice their sum,
and the area/perimeter denominator has floor $10^{-12}$. Exact IoU-one
entries and the anchor's own entry are excluded from the last four IoU
statistics, but remain in the other neighborhood summaries. Empty positive
overlap sets give four zeros. Reduction is float64, the 40 statistics
are stored float32, and the complete input is float32. MetaStats141
appends Full100's native-set columns 97--99.

This implementation uses standard pixel area/perimeter in place of the
legacy author-code geometry expressions. It also uses the common
pre-NMS bank, eligible/ignore target, HGB and source OP. The original
MetaDetect experiments concern detection quality estimates and have
different targets and learners; their feature dimensionality is not
universally 138. These differences distinguish a matched task adaptation
from a numerical reproduction of a published system.

The detection MetaFusion-inspired variant follows meta-score replacement
after raw selection~\cite{riedlinger2023gradient}. Raw Hard-NMS at
IoU 0.5 operates on the common score-0.005 bank without an intermediate
survivor cap. Survivors receive their calibrated meta score; their
independently selected threshold and a final cap of 300 are then applied.
The trained head is reused. This variant includes no gradient features,
Monte Carlo passes or coordinate regressor. Since its raw-NMS setting
and selected threshold also differ from pre-NMS Hard-0.4 scoring,
the comparison is between complete policies, not a one-factor estimate
of the causal effect of operation order.

\begin{table}[tbp]
\centering\footnotesize
\caption{Image cohorts and analysis roles. FIT/OP/DEV contain source-task records. CH cohorts are distinct same-source confirmations; MIAP and MOT17Det test unchanged source-fitted policies. Timing-128 is a CH-255/MIAP-256 subset, and Timing-64 is a CH-319/MOT subset.}\label{tab:data}
\adjustbox{width=1\linewidth}{
\begin{tabular}{lrrl}
\toprule
Role/cohort & Images & Eligible GT & Use \\
\midrule
FIT & 1024 & -- & Readout/projection fitting \\
OP & 512 & 11428 & Calibration and policy selection \\
DEV & 256 & 5668 & Adaptive development \\
CH-255 & 255 & 5332 & Source confirmation \\
MIAP-256 & 256 & 1119 & Source-frozen transfer \\
Info-320 & 320 & 7278 & Information-group confirmation \\
Timing-128 & 128 & 1513 & 64 CH + 64 MIAP; runtime \\
CH-319 & 319 & 7246 & Direct-reference confirmation \\
MOT17Det & 223 & 4491 & External frames; seven sequences \\
Timing-64 & 64 & 1253 & 32 CH-319 + 32 MOT; runtime \\
\bottomrule
\end{tabular}
}
\end{table}

\begin{table}[!t]\centering\small
\caption{Direct output-statistics controls on DEV-256, at seed 25 and the primary source-OP budget. New readouts share FIT rows, labels, HGB settings and OP opportunities. MetaStats variants adapt published feature families; post-NMS variants replace scores after raw NMS .5, whereas the other learned rows use meta-score Hard-NMS .4. AP50 uses the author-visible protocol.}\label{tab:directdev}
\adjustbox{width=1\linewidth}{
\begin{tabular}{lrrrr}\toprule
Policy & Dim. & TP & UP & AP50 (\%) \\ \midrule
Native & -- & 4338 & 1068 & 72.89 \\
Raw Hard .4 & -- & 4438 & 1105 & 74.52 \\
Raw OP-best WBF & -- & 4475 & 1093 & 75.49 \\
Full100 & 100 & 4591 & 1141 & 78.34 \\
Score2Native3 & 5 & 4455 & 1099 & 74.96 \\
MetaStats & 138 & 4578 & 1120 & 78.17 \\
MetaStats+Native3 & 141 & 4585 & 1107 & 78.33 \\
Post-NMS meta score & 138 & 4341 & 1111 & 73.92 \\
Post-NMS meta score & 141 & 4445 & 1077 & 76.00 \\
\bottomrule\end{tabular}
}
\end{table}

\begin{table}[tbp]
\centering\small
\caption{Full100 minus the prespecified information references on Info-320. Individual 98.75\% paired-image bootstrap intervals comprise a four-endpoint family with an approximate 95\% coverage target. Positive UP differences indicate greater prediction burden.}\label{tab:infoconfirmintervals}
\adjustbox{width=1\linewidth}{
\begin{tabular}{lrrr}
\toprule
Reference & Endpoint & Difference & Interval \\
\midrule
Score2 & Recall (pp) & +3.119 & [+2.175, +4.053] \\
Score2 & UP/image & +0.25312 & [-0.01562, +0.53125] \\
ClassGeom97 & Recall (pp) & +3.779 & [+2.739, +4.857] \\
ClassGeom97 & UP/image & +0.30000 & [+0.00625, +0.59688] \\
\bottomrule
\end{tabular}
}
\end{table}

\subsection{ROI and graph/quality controls}
ROI132 adds 32 projected same-forward appearance entries to Full100.
The box-head vector $h\in\mathbb R^{1024}$ is stored as float16 and
converted to float32 for projection. PCA uses up to 100000 unlabeled
FIT candidates independently of the classifier sample. The transform is
\begin{align}
u&=h/\max(\|h\|_2,10^{-8}),\\
z&=\operatorname{clip}\big(D^{-1}V(u-\mu),-8,8\big).
\end{align}
PCA has 32 components, randomized solver, seed 25 and four power
iterations. The scale is the population standard deviation of each
actual FIT \texttt{fit\_transform} component, floored at $10^{-6}$.
The saved mean, components and scales are float32. Export, transfer,
projection and the readout are inference dependencies even though
there is no second image encoder.

The self-built graph scorer is a task adaptation of the
Learning-NMS/GossipNet architecture~\cite{hosang2017nms}, combined
with a separate candidate-validity score. Eight residual gossip blocks
use latent dimension 128, reduced dimension 32 and pair-hidden
dimension 64. Nodes use Full132, with neighbors at IoU $\geq0.2$
including self. Nine pair inputs encode upstream scores, IoU, relative
center displacement/distance, log-base-2 width/height ratios and their
difference. Pair inputs are clipped to $[-20,20]$ and messages are
max-aggregated. The output logit adds the clipped upstream score's logit.
The graph uses matching-dependent one-to-one targets, whereas the
independent validity model permits duplicate positives. Fusion is
$\sqrt{gv}$, with threshold 0.26291174000257606, Hard-NMS 0.5 and
cap 300. It pays for upstream score $q$, graph $g$ and validity $v$,
sharing the projection once per request. Its supervision and dependencies
make it a stronger-resource policy, not a unary-feature ablation.

\begin{table*}[!t]
\centering\footnotesize
\caption{All information-group results on DEV-256. Cells report TP/UP/policy AP50 (\%). Budgets are multiples of source-OP Native UP=2388 and apply to OP selection. Seed 25 is primary; seeds 26/27 assess training variation.}\label{tab:informationbudgets}
\begin{tabular}{lrrrr}
\toprule
Seed & Input & $0.75B$ & $B$ & $1.25B$ \\
\midrule
25 & Score2 & 4226/827/71.39 & 4422/1084/74.23 & 4540/1341/75.87 \\
25 & Class91 & 4154/808/70.99 & 4343/1129/73.70 & 4455/1430/75.24 \\
25 & ClassGeom97 & 4209/811/71.93 & 4375/1089/74.33 & 4485/1388/75.86 \\
25 & Full100 & 4488/845/76.85 & 4591/1141/78.34 & 4648/1431/79.13 \\
26 & Score2 & 4198/803/70.95 & 4422/1088/74.20 & 4549/1390/75.96 \\
26 & Class91 & 4169/842/71.23 & 4335/1112/73.62 & 4448/1435/75.18 \\
26 & ClassGeom97 & 4214/816/72.03 & 4376/1099/74.37 & 4490/1377/75.94 \\
26 & Full100 & 4474/825/76.67 & 4578/1142/78.18 & 4644/1400/79.10 \\
27 & Score2 & 4236/840/71.53 & 4418/1088/74.16 & 4544/1397/75.91 \\
27 & Class91 & 4170/819/71.28 & 4337/1107/73.69 & 4443/1418/75.15 \\
27 & ClassGeom97 & 4225/822/72.20 & 4375/1077/74.36 & 4484/1396/75.88 \\
27 & Full100 & 4467/831/76.63 & 4590/1113/78.42 & 4650/1422/79.25 \\
\bottomrule
\end{tabular}
\end{table*}

\begin{table*}[!t]\centering\footnotesize
\caption{Complete new-reference DEV results. Each cell is TP/UP/policy AP50 (\%). Budgets are $0.75B,B,1.25B$ with $B=2388$ source-OP UP. Seed 25 is primary; seeds 26/27 describe training variation. These are the same reused 256 development images.}\label{tab:directbudgets}
\setlength{\tabcolsep}{4pt}
\begin{tabular}{rlrrr}\toprule
Seed & Readout & $0.75B$ & $B$ & $1.25B$ \\ \midrule
25 & Score2Native3 & 4274/827/72.33 & 4455/1099/74.96 & 4569/1375/76.54\\
25 & MetaStats138 & 4454/805/76.36 & 4578/1120/78.17 & 4645/1412/79.10\\
25 & MetaStats141 & 4469/815/76.64 & 4585/1107/78.33 & 4642/1393/79.12\\
25 & Post-NMS 138 & 4224/835/72.24 & 4341/1111/73.92 & 4447/1396/75.37\\
25 & Post-NMS 141 & 4334/804/74.39 & 4445/1077/76.00 & 4539/1373/77.30\\
\midrule
26 & Score2Native3 & 4254/827/72.00 & 4449/1107/74.82 & 4552/1397/76.25\\
26 & MetaStats138 & 4465/854/76.52 & 4567/1125/78.00 & 4641/1384/79.02\\
26 & MetaStats141 & 4462/844/76.50 & 4584/1151/78.27 & 4640/1417/79.05\\
26 & Post-NMS 138 & 4223/829/72.26 & 4342/1104/73.98 & 4449/1403/75.45\\
26 & Post-NMS 141 & 4334/814/74.37 & 4437/1104/75.87 & 4531/1366/77.17\\
\midrule
27 & Score2Native3 & 4261/833/72.06 & 4455/1093/74.87 & 4556/1385/76.27\\
27 & MetaStats138 & 4446/830/76.30 & 4580/1109/78.26 & 4643/1431/79.13\\
27 & MetaStats141 & 4476/822/76.80 & 4594/1110/78.52 & 4649/1391/79.29\\
27 & Post-NMS 138 & 4216/821/72.16 & 4338/1105/73.92 & 4449/1392/75.45\\
27 & Post-NMS 141 & 4334/830/74.33 & 4450/1067/76.02 & 4520/1352/77.00\\
\bottomrule\end{tabular}
\end{table*}

\subsection{Separately fitted RetinaNet interface}
RetinaNet R50-FPNv2 COCO\_V1~\cite{lin2017focal,torchvision_retina_v2}
uses short side 1000 and maximum side 1667. Native output retains
the top 1000 class-pair candidates per feature-pyramid level, score
above 0.01, classwise NMS 0.5 and total cap 300, then person score
at least 0.4. The released person bank uses score at least 0.005 and
stable global raw-score top 1000 across levels. Its 91 sigmoid channels
have different semantics from the primary network's softmax background.

A separate 100-feature HGB uses RetinaNet FIT1024 candidates and the
same principal HGB/sampling settings. Its OP512 UP ceiling is 1054.
Quality-WBF uses only this head, with no extra Platt map, ROI or primary
detector head. Raw Gaussian uses sigma 0.5 and threshold
0.30792319303596327. Raw WBF uses IoU 0.4, maximum confidence and
threshold 0.33600887656211853; quality WBF uses the same operator
and threshold 0.7635001540184021. All outputs are capped at 300.

\begin{table}[tbp]
\centering\small
\caption{Object-set changes on Info-320, Full100 relative to each reference. $G$ counts annotations matched only by Full100, and $L$ those matched only by the reference. Net changes are Full100 minus reference.}\label{tab:infosets}
\begin{tabular}{lrrrr}
\toprule
Reference & $G$ & $L$ & $\Delta$TP & $\Delta$UP \\
\midrule
Native & 405 & 119 & +286 & +40 \\
Raw Hard .4 & 328 & 129 & +199 & +41 \\
Raw OP-best WBF & 287 & 149 & +138 & +61 \\
Score2 & 347 & 120 & +227 & +81 \\
Class91 & 412 & 126 & +286 & +80 \\
ClassGeom97 & 382 & 107 & +275 & +96 \\
\bottomrule
\end{tabular}
\end{table}

\begin{table}[tbp]
\centering\small
\caption{Output accounting on Info-320: outputs = TP + UP + ignored. Ignored predictions are separate from unmatched predictions.}\label{tab:infooutputs}
\begin{tabular}{lrrrr}
\toprule
Policy & TP & UP & Ignored & Outputs \\
\midrule
Native & 5708 & 1464 & 1458 & 8630 \\
Raw Hard .4 & 5795 & 1463 & 1694 & 8952 \\
Raw OP-best WBF & 5856 & 1443 & 1797 & 9096 \\
Score2 & 5767 & 1423 & 1648 & 8838 \\
Class91 & 5708 & 1424 & 1242 & 8374 \\
ClassGeom97 & 5719 & 1408 & 1207 & 8334 \\
Full100 & 5994 & 1504 & 1942 & 9440 \\
\bottomrule
\end{tabular}
\end{table}

\section{Operating Points, Matching and Statistics}
\subsection{Scalar maps and output rules}
SHA-256 image-key parity defines two OP folds. Each scalar logistic
fit uses $C=1$, at most 300 iterations and an unconstrained slope.
The opposite fold receives its out-of-fold map, while a full-OP map
is saved for deployment. Fold-specific maps need not induce a common
pooled ranking. A common strictly increasing map would preserve a fixed
Hard-NMS ordering when its threshold is transformed consistently;
that observation alone does not cover this OOF selection procedure,
clipping ties or coordinate-changing WBF.

Thresholds maximize OP TP under the chosen UP ceiling. Within an
operator, ties prefer lower UP, fewer Native-relative object losses
and a higher threshold; recipe name breaks the last cross-operator tie.
Equal-score events are atomic. At the principal ceiling of 2388,
Score2/Class91/ClassGeom97/Full100 realize 2376/2386/2387/2388 OP UP.
Separate budget multipliers are 0.75 and 1.25. These are source
selection opportunities, not constraints imposed on confirmation UP.

Raw selection considers Hard-NMS IoU 0.3/0.4/0.5/0.6, Gaussian
Soft-NMS sigma 0.25/0.5/1, and WBF IoU 0.4/0.55/0.7. The matched
information and direct meta-classification comparisons use Hard-0.4
throughout. Representation sensitivity additionally examines Gaussian
0.5 and WBF 0.4. The raw references thus receive broader operator
selection rather than being restricted to a deliberately weak default.

Hard inference clips coordinates, removes invalid/non-finite rows and
scores below threshold, and applies stable score ordering. It suppresses
at IoU strictly greater than the configured threshold, retaining equality.
OP frontier construction starts from fixed floor-zero operator outputs
with cap 300 and scans global score-prefix events, rematching selected
outputs. The deployed Hard implementation remains threshold-first.
Single-model WBF~\cite{solovyev2021wbf} clusters valid floor-zero inputs
with maximum confidence, then applies the output threshold and stable
cap. Learned scores can alter clustering, coordinates and work, so its
cost is not assumed equal to raw-score WBF.

\subsection{Count and AP semantics}
For CrowdHuman counts, valid, non-ignored person visible boxes are
eligible; other annotation rows with valid full boxes form ignore
regions. Score-descending greedy matching uses unused eligible objects
at IoU $\geq0.5$, followed by ignore tests for unmatched predictions
at intersection-over-prediction-area $\geq0.5$. Empty-GT images remain
in image-level analyses. The object sets are actual eligible annotation
identities assigned by that matcher.

CrowdHuman author-visible AP replaces all full boxes by visible boxes
while preserving ignore/tag values. Author code clips pixel boundaries,
uses strict IoU $>0.5$ and strict ignore overlap $>0.5$, and excludes
ignored predictions from ranked events. AP sums consecutive trapezoids
starting at the second event. Thus its TP/UP flags need not be identical
to the count matcher at boundaries or ignored geometry.

MIAP uses native person boxes; group-of, depiction-of and inside-of rows
are ignored, while occluded and truncated individuals remain eligible.
Age/gender attributes are unused. These annotations are not evidence of
exhaustive physical-person coverage. MOT17Det uses active class-1 full
boxes as eligible; classes 2, 7, 8, 12 or non-positive active marks
define ignore boxes. Other classes are not targets. One is subtracted
from top-left coordinates to convert one-based annotation positions.
No visibility cutoff is imposed. Count matching retains eligible-first
IoU/IoA $\geq0.5$. MIAP/MOT policy AP globally sorts native-box
events, excludes ignored outputs and integrates the monotone all-point
precision envelope. This MOT adaptation is not its official leaderboard
metric. All AP values use the deployed thresholded/capped output.

\subsection{Resampling families}
For paired resampled images $S$, recall and UP differences are
\begin{align}
\Delta R(S)&=\frac{\sum_{i\in S}(\mathrm{TP}_{i,a}-\mathrm{TP}_{i,r})}
{\sum_{i\in S}|G_i|},\\
\Delta U(S)&=\frac{\sum_{i\in S}(\mathrm{UP}_{i,a}-\mathrm{UP}_{i,r})}{|S|}.
\end{align}
Image multiplicities affect numerator and denominator. The CH-255/MIAP
family compares Full100 with source-selected raw WBF using 20000
paired draws and PCG64(2807), reset per source. Its four endpoints
receive individual 98.75\% linear-percentile intervals. Info-320 uses
50000 paired draws with PCG64(2917) for Full100 versus Score2 and
ClassGeom97, each on recall and UP/image. Each family has its own
approximate 95\% Bonferroni coverage target; neither is pooled with
later families.

\begin{table}[tbp]
\centering\small
\caption{Source-task and zero-tuning transfer results. Full100 and ROI use seed 25 and Hard-NMS .4; $g/v$ uses Hard-NMS .5. $G/L$ are relative to raw OP-selected WBF. AP50 follows the CrowdHuman author-visible or MIAP native-person protocol. MIAP uses unchanged source-fitted heads and thresholds.}\label{tab:confirmationH}
\adjustbox{width=1\linewidth}{
\begin{tabular}{lrrrrr}
\toprule
Policy & TP & UP & $G$ & $L$ & AP50 (\%) \\
\midrule
\multicolumn{6}{l}{CrowdHuman (255 images, 5332 eligible objects)} \\
Native & 4310 & 1073 & 122 & 183 & 76.73 \\
Raw Hard .4 & 4366 & 1069 & 101 & 106 & 77.96 \\
Raw OP-best WBF & 4371 & 1057 & 0 & 0 & 78.45 \\
Full100 & 4461 & 1071 & 193 & 103 & 81.15 \\
ROI readout & 4472 & 1065 & 223 & 122 & 81.67 \\
$g/v$ fusion & 4482 & 1072 & 223 & 112 & 81.94 \\
\multicolumn{6}{l}{MIAP (256 images, 1119 eligible objects)} \\
Native & 942 & 787 & 34 & 26 & 77.12 \\
Raw Hard .4 & 938 & 809 & 28 & 24 & 76.90 \\
Raw OP-best WBF & 934 & 839 & 0 & 0 & 77.18 \\
Full100 & 933 & 845 & 31 & 32 & 76.34 \\
ROI readout & 919 & 854 & 27 & 42 & 75.27 \\
$g/v$ fusion & 932 & 834 & 38 & 40 & 76.15 \\
\bottomrule
\end{tabular}
}
\end{table}

\begin{table}[tbp]
\centering\footnotesize
\caption{Full100 minus raw OP-selected WBF in the source/transfer confirmation. Individual 98.75\% paired-image intervals comprise a four-endpoint family with an approximate 95\% coverage target.}\label{tab:primary}
\adjustbox{width=1\linewidth}{
\begin{tabular}{lrrrr}
\toprule
Source & $\Delta$ recall (pp) & Interval & $\Delta$UP/image & Interval \\
\midrule
CrowdHuman & +1.688 & [+0.773, +2.598] & +0.055 & [-0.224, +0.341] \\
MIAP & -0.089 & [-1.903, +1.579] & +0.023 & [-0.266, +0.379] \\
\bottomrule
\end{tabular}
}
\end{table}

\begin{table}[tbp]
\centering\small
\caption{Source-task and zero-tuning transfer results for the separately fitted RetinaNet interface. The adapted policies use Gaussian $\sigma=.5$ or WBF .4; $G/L$ are relative to raw Gaussian. AP50 follows the CrowdHuman author-visible or MIAP native-person protocol. Source-fitted heads and thresholds are unchanged on MIAP.}\label{tab:confirmationRetina}
\begin{tabular}{lrrrrr}
\toprule
Policy & TP & UP & $G$ & $L$ & AP50 (\%) \\
\midrule
\multicolumn{6}{l}{CrowdHuman (255 images, 5332 eligible objects)} \\
Native & 3229 & 534 & 74 & 365 & 58.35 \\
Raw Gaussian & 3520 & 507 & 0 & 0 & 63.68 \\
Raw WBF & 3489 & 486 & 137 & 168 & 63.29 \\
Validity WBF & 3691 & 566 & 318 & 147 & 67.47 \\
\multicolumn{6}{l}{MIAP (256 images, 1119 eligible objects)} \\
Native & 802 & 298 & 14 & 29 & 67.84 \\
Raw Gaussian & 817 & 361 & 0 & 0 & 69.00 \\
Raw WBF & 819 & 352 & 26 & 24 & 69.21 \\
Validity WBF & 831 & 336 & 44 & 30 & 69.92 \\
\bottomrule
\end{tabular}
\end{table}

\begin{table}[tbp]
\centering\footnotesize
\caption{Annotation sensitivity on the fixed CH-255 predictions. Each column uses the CrowdHuman author evaluator with the indicated box semantics.}\label{tab:boxsemantics}
\adjustbox{width=1\linewidth}{
\begin{tabular}{lrr}
\toprule
Policy & vbox policy AP50 (\%) & fbox policy AP50 (\%) \\
\midrule
H: Native & 76.73 & 57.63 \\
H: Raw Hard .4 & 77.96 & 57.59 \\
H: Raw OP-best WBF & 78.45 & 57.54 \\
H: Full100 & 81.15 & 63.63 \\
H: ROI readout & 81.67 & 63.85 \\
H: $g/v$ fusion & 81.94 & 64.31 \\
Retina: Native & 58.35 & 47.90 \\
Retina: Raw Gaussian & 63.68 & 50.12 \\
Retina: Raw WBF & 63.29 & 49.87 \\
Retina: Validity WBF & 67.47 & 55.98 \\
\bottomrule
\end{tabular}
}
\end{table}

\begin{table*}[tbp]\centering\footnotesize
\renewcommand{\arraystretch}{0.94}
\caption{Complete quality and output accounting for the 12 fixed policies in each new cohort. FN counts unmatched eligible annotations; ignored output is separate from UP. AP50 follows the source-specific protocols in S2. All reviewer-call counts are zero.}\label{tab:newaccounting}
\setlength{\tabcolsep}{4pt}
\begin{tabular}{lrrrrrr}\toprule
Policy & TP & UP & FN & Outputs & Ignored & AP50 (\%) \\ \midrule
\multicolumn{7}{l}{CrowdHuman, 319 images} \\
Native & 5684 & 1342 & 1562 & 8636 & 1610 & 74.19 \\
Raw Hard .4 & 5802 & 1348 & 1444 & 9060 & 1910 & 75.92 \\
Raw OP-best WBF & 5848 & 1387 & 1398 & 9225 & 1990 & 76.77 \\
Full100 & 6004 & 1366 & 1242 & 9559 & 2189 & 80.10 \\
Score2 & 5782 & 1319 & 1464 & 8943 & 1842 & 75.61 \\
Class91 & 5694 & 1384 & 1552 & 8481 & 1403 & 75.37 \\
ClassGeom97 & 5730 & 1416 & 1516 & 8537 & 1391 & 75.64 \\
Score2Native3 & 5804 & 1385 & 1442 & 9191 & 2002 & 76.05 \\
MetaStats138 & 6002 & 1392 & 1244 & 9485 & 2091 & 80.16 \\
MetaStats141 & 5990 & 1387 & 1256 & 9527 & 2150 & 80.00 \\
Post-NMS 138 & 5719 & 1406 & 1527 & 8595 & 1470 & 76.22 \\
Post-NMS 141 & 5856 & 1389 & 1390 & 8936 & 1691 & 78.27 \\
\midrule
\multicolumn{7}{l}{MOT17Det, 223 frames} \\
Native & 2981 & 1350 & 1510 & 5018 & 687 & 61.26 \\
Raw Hard .4 & 2994 & 1492 & 1497 & 5222 & 736 & 61.45 \\
Raw OP-best WBF & 2988 & 1570 & 1503 & 5331 & 773 & 61.34 \\
Full100 & 3030 & 1349 & 1461 & 5094 & 715 & 63.71 \\
Score2 & 2984 & 1478 & 1507 & 5203 & 741 & 61.16 \\
Class91 & 2947 & 1159 & 1544 & 4740 & 634 & 61.92 \\
ClassGeom97 & 2962 & 1127 & 1529 & 4657 & 568 & 62.33 \\
Score2Native3 & 2973 & 1456 & 1518 & 5184 & 755 & 60.80 \\
MetaStats138 & 3000 & 1334 & 1491 & 5033 & 699 & 62.71 \\
MetaStats141 & 3010 & 1348 & 1481 & 5069 & 711 & 63.18 \\
Post-NMS 138 & 2910 & 1128 & 1581 & 4615 & 577 & 61.08 \\
Post-NMS 141 & 2943 & 1163 & 1548 & 4701 & 595 & 61.80 \\
\bottomrule\end{tabular}
\end{table*}

\begin{table}[tbp]\centering\small
\caption{MOT17Det sequence-level changes, Full100 minus reference. Cells are $\Delta$TP/$\Delta$UP. The seven videos are the block units used for the approximate primary analysis.}\label{tab:motsequences}
\adjustbox{width=1\linewidth}{
\begin{tabular}{lrrrr}\toprule
Sequence & Frames / GT & Score2Native3 & MetaStats141 & Native \\ \midrule
MOT17-02 & 32/990 & +29/+19 & +12/+4 & +36/+44\\
MOT17-04 & 32/1449 & +7/-7 & +11/-26 & +5/-19\\
MOT17-05 & 31/263 & +6/-24 & +1/+22 & +4/-16\\
MOT17-09 & 32/327 & +0/-21 & +1/+4 & +1/-17\\
MOT17-10 & 32/626 & +6/-7 & -4/+2 & +4/+9\\
MOT17-11 & 32/336 & +4/-44 & +1/+3 & +0/-23\\
MOT17-13 & 32/500 & +5/-23 & -2/-8 & -1/+21\\
\bottomrule\end{tabular}
}
\end{table}

\begin{table}[tbp]\centering\small
\caption{Complete warm requests on the same fixed 64 images (32 CrowdHuman and 32 MOT17Det). Each job averages three real requests within image before aggregation. The two independent jobs reverse policy order; their means and 95th percentiles are reported separately in milliseconds. C99 changes only the execution of frozen MetaStats features. These are engineering workload distributions, not external-population intervals.}\label{tab:optimizedcost}
\begin{tabular}{lrrrr}\toprule
 & \multicolumn{2}{c}{Job 1} & \multicolumn{2}{c}{Job 2} \\
Policy & Mean & p95 & Mean & p95 \\\midrule
Raw Hard .4 & 89.28 & 102.12 & 89.04 & 101.72 \\
Full100 & 95.37 & 108.84 & 94.77 & 105.94 \\
Score2Native3 & 94.22 & 111.40 & 92.95 & 103.02 \\
Meta138 / NumPy & 177.64 & 228.33 & 176.48 & 226.08 \\
Meta138 / C99 & 95.21 & 108.79 & 95.04 & 106.88 \\
Meta141 / NumPy & 182.30 & 232.56 & 177.68 & 230.15 \\
Meta141 / C99 & 96.78 & 113.82 & 95.53 & 106.60 \\
\bottomrule\end{tabular}
\end{table}

\begin{table*}[tbp]\centering\small
\caption{Sourcewise complete-request distributions for job 1 of the exact implementation comparison, in milliseconds. Each entry uses 32 image means of three requests. Repeat SD is the mean of within-image sample standard deviations.}\label{tab:optimizedcostsources}
\begin{tabular}{llrrrr}\toprule
Job / source & Policy & Mean & Median & p95 & Repeat SD \\\midrule
1 / CH & Raw Hard .4 & 89.39 & 82.42 & 136.07 & 1.78 \\
 & Full100 & 95.31 & 88.10 & 140.80 & 2.04 \\
 & Score2Native3 & 94.92 & 86.95 & 151.78 & 4.42 \\
 & Meta138 / NumPy & 177.92 & 174.58 & 230.95 & 3.42 \\
 & Meta138 / C99 & 94.88 & 87.82 & 141.12 & 1.73 \\
 & Meta141 / NumPy & 179.88 & 172.99 & 235.07 & 3.94 \\
 & Meta141 / C99 & 97.57 & 89.01 & 150.53 & 4.41 \\
\midrule
1 / MOT & Raw Hard .4 & 89.16 & 92.04 & 94.05 & 1.53 \\
 & Full100 & 95.44 & 98.23 & 101.56 & 2.08 \\
 & Score2Native3 & 93.52 & 95.96 & 100.55 & 2.09 \\
 & Meta138 / NumPy & 177.35 & 174.44 & 217.94 & 2.19 \\
 & Meta138 / C99 & 95.54 & 98.37 & 101.26 & 2.20 \\
 & Meta141 / NumPy & 184.71 & 176.60 & 225.89 & 11.62 \\
 & Meta141 / C99 & 95.99 & 98.91 & 101.36 & 1.79 \\
\bottomrule\end{tabular}
\end{table*}

\begin{table*}[tbp]\centering\small
\caption{Sourcewise complete-request distributions for job 2 of the exact implementation comparison, in milliseconds. Each entry uses 32 image means of three requests. Repeat SD is the mean of within-image sample standard deviations.}\label{tab:optimizedcostsourcessecond}
\begin{tabular}{llrrrr}\toprule
Job / source & Policy & Mean & Median & p95 & Repeat SD \\\midrule
2 / CH & Raw Hard .4 & 88.91 & 81.91 & 136.33 & 1.23 \\
 & Full100 & 94.92 & 88.12 & 144.63 & 1.52 \\
 & Score2Native3 & 92.88 & 85.99 & 140.72 & 1.24 \\
 & Meta138 / NumPy & 176.84 & 173.01 & 231.18 & 4.17 \\
 & Meta138 / C99 & 95.48 & 88.62 & 146.32 & 1.75 \\
 & Meta141 / NumPy & 177.76 & 172.46 & 231.77 & 2.50 \\
 & Meta141 / C99 & 95.77 & 88.75 & 142.85 & 1.39 \\
\midrule
2 / MOT & Raw Hard .4 & 89.18 & 92.45 & 94.53 & 1.20 \\
 & Full100 & 94.61 & 97.56 & 100.25 & 1.31 \\
 & Score2Native3 & 93.02 & 95.86 & 98.52 & 1.13 \\
 & Meta138 / NumPy & 176.12 & 174.22 & 215.40 & 1.86 \\
 & Meta138 / C99 & 94.59 & 97.73 & 99.54 & 1.11 \\
 & Meta141 / NumPy & 177.60 & 174.13 & 217.98 & 1.95 \\
 & Meta141 / C99 & 95.29 & 98.29 & 101.06 & 1.10 \\
\bottomrule\end{tabular}
\end{table*}

\begin{table}[tbp]\centering\small
\caption{Mean request components on the balanced fixed 64-image workload, in milliseconds. Forward time includes the required capture path; feature time precedes the combined statistics, readout, scalar mapping and output operator. Components are measured within complete requests and are not separate microbenchmark totals.}\label{tab:optimizedcomponents}
\adjustbox{width=1\linewidth}{
\begin{tabular}{llrrr}\toprule
Job & Policy & Forward & Base features & Readout / output \\\midrule
1 & Raw Hard .4 & 79.42 & 0.09 & 0.90 \\
 & Full100 & 80.61 & 1.27 & 4.71 \\
 & Score2Native3 & 80.48 & 0.84 & 4.05 \\
 & Meta138 / NumPy & 79.49 & 0.28 & 89.06 \\
 & Meta138 / C99 & 79.41 & 0.28 & 6.76 \\
 & Meta141 / NumPy & 80.52 & 1.16 & 91.68 \\
 & Meta141 / C99 & 80.47 & 1.17 & 6.31 \\
\midrule
2 & Raw Hard .4 & 79.25 & 0.10 & 0.90 \\
 & Full100 & 80.27 & 1.27 & 4.42 \\
 & Score2Native3 & 80.10 & 0.83 & 3.23 \\
 & Meta138 / NumPy & 79.46 & 0.28 & 87.92 \\
 & Meta138 / C99 & 79.54 & 0.29 & 6.35 \\
 & Meta141 / NumPy & 80.22 & 1.16 & 87.41 \\
 & Meta141 / C99 & 80.10 & 1.15 & 5.47 \\
\bottomrule\end{tabular}
}
\end{table}

CH-319 uses 30000 paired-image draws with PCG64(3117), and MOT
uses 30000 sequence-block draws with PCG64(3118). Each has the two
Full100 contrasts to Score5 and MetaStats141 and the same two endpoints.
All frames in a resampled sequence receive its multiplicity, and pooled
recall and UP/frame are recomputed with their GT/frame totals. RNG state
is reset per pair to give common draws within a source. There are no
zero-total-GT draws. The seven-sequence approximation has limited
population precision. Development intervals are pointwise descriptions
of adaptive results. AP, extra references and G/L remain secondary;
no equivalence or non-inferiority conclusion follows from intervals
including zero. This study does not transfer a historical risk certificate
to the present readouts.

\section{Cohorts and Additional Results}

\subsection{Record roles and source screening}
The source images come from the archived official CrowdHuman training
source~\cite{shao2018crowdhuman}, not from a newly untouched official
test split. FIT, OP and DEV support their stated adaptive roles.
CH-255 and MIAP-256 evaluate fixed representation policies; Info-320
contains distinct same-source records for the matched information
comparison. Info planning used development influence-quantity variance
and normal-approximation half-width targets of 1.5 recall percentage
points and 0.5 UP/image. The largest estimate was 297 records, rounded
to 320. This is precision planning, not a guarantee of power or success.

Direct-reference confirmation selected 320 new source records and 224
MOT17Det frames, initially 32 evenly spaced frames per training video.
Identity, byte/RGB hash and available perceptual/metadata checks screened
earlier roles before label interpretation. One record per cohort was
excluded without replacement, leaving 319 and 223. All twelve frozen
policies were evaluated on each retained record, and both cohorts'
predictions were completed before either annotation evaluation. This
supports record-disjoint prospective evaluation; unknown scene and
pretraining overlap remain possible.

The model and source policies are unchanged on MIAP/Open
Images~\cite{schumann2021miap,kuznetsova2020openimages} and MOT17Det.
Timing subsets are computational workloads within the consumed cohorts,
not additional independent quality tests. Their labels were closed for
the implementation optimization, and no policy was selected from them.

\subsection{Information and direct-reference development}
Table~\ref{tab:directdev} retains the main-seed direct controls.
The following paired seed/budget tables show the selected TP, UP and
policy AP50, including points unfavorable to Full100. They use one
adaptive DEV cohort and therefore do not provide independent replication
for every cell.

\subsection{Confirmation and output accounting}
The Info-320 primary intervals are in Table~\ref{tab:infoconfirmintervals};
Tables~\ref{tab:infosets} and \ref{tab:infooutputs} separate matched-object
changes and ignored outputs. Counts with the same total can refer to
different identities: Class91 and Native each match 5708 objects but
exchange 287 in each direction. Such identities are matcher-dependent
descriptions, not an additional ground-truth discovery model.

The CH-255/MIAP representation controls, their original primary intervals,
and the separately fitted RetinaNet are retained below. RetinaNet's
positive MIAP coordinate belongs to that detector/readout policy; it
cannot replace the two-stage Full100 transfer result. Visible/full-box
sensitivity is kept separate because changing box semantics changes the
evaluation target.

Table~\ref{tab:newaccounting} includes all twelve direct-reference
policies, not just the selected main-text rows. Table~\ref{tab:motsequences}
shows the seven temporal units used for block resampling.

\section{Complete Runtime and Execution Equivalence}
\subsection{Measurement boundary}
Complete requests use an NVIDIA A800-SXM4-80GB, Python 3.11.17,
PyTorch 2.6.0+cu124, TorchVision 0.21.0+cu124, NumPy 2.4.6,
scikit-learn 1.9.1 and Pillow 12.3.0. Batch size is one, with two
CPU worker threads. Detector inference is FP32, highest matmul precision
and TF32 disabled. Each policy performs image read/byte verification,
RGB decoding, its own forward, required capture/transfers, feature/head
processing and final output, then synchronizes. Unused inputs are omitted:
raw does not pay for native multiclass NMS or full class-probability
transfer; Meta138 also omits native processing; Score5 transfers only
two probability slots plus native boxes/scores.

Warmup, loading, compilation, fitting and adaptation data preparation
are outside this clock. Nested components are diagnostic and are not
added to the complete time. Actual policy outputs on the timed images
are checked, so a cached feature bank cannot substitute for a measured
forward. The GPU has other idle resident contexts and is resource-guarded,
not exclusive hardware. Neither future service latency nor whole-batch
economic break-even follows from these warm-request means.

\subsection{Equivalent statistics execution}
The reference implementation constructs a dense float64 IoU matrix and
per-anchor neighborhood reductions. Its C99 replacement uses linear numeric
scratch storage and caches shared reductions only for exactly identical
neighborhood memberships. Anchor geometry and overlap summaries remain
separate, and the cache lasts for one feature call. All pairwise candidate
tests remain; this is an execution optimization rather than a linear-time
selection method.

\begin{table}[tbp]
\centering\small
\caption{Same-Hard timing session on 64 CrowdHuman and 64 MIAP images. Each image has three interleaved requests per policy; timing statistics use their within-image means. Both policies use Hard-NMS .4. TP/UP describe these same 128 images.}\label{tab:costsamehard}
\adjustbox{width=1\linewidth}{
\begin{tabular}{lrrrr}
\toprule
Source & Policy & TP & UP & Mean ms \\
\midrule
CrowdHuman & Raw Hard .4 & 1051 & 236 & 87.317 \\
CrowdHuman & Full100 Hard .4 & 1068 & 245 & 91.914 \\
MIAP & Raw Hard .4 & 240 & 247 & 78.712 \\
MIAP & Full100 Hard .4 & 247 & 270 & 82.164 \\
Fixed 64/64 mix & Raw Hard .4 & 1291 & 483 & 83.015 \\
Fixed 64/64 mix & Full100 Hard .4 & 1315 & 515 & 87.039 \\
\bottomrule
\end{tabular}
}
\end{table}

\begin{table}[tbp]
\centering\small
\caption{Matched-WBF RetinaNet timing session on 64 images per source, with three interleaved repetitions. TP/UP describe these timed outputs. Policies were unchanged; this session followed disclosure of the quality results.}\label{tab:costmatched}
\begin{tabular}{lrrr}
\toprule
Policy & TP & UP & Mean ms \\
\midrule
\multicolumn{4}{l}{CrowdHuman (64 images)} \\
Raw WBF & 871 & 108 & 175.58 \\
Validity WBF & 890 & 104 & 188.45 \\
\multicolumn{4}{l}{MIAP (64 images)} \\
Raw WBF & 202 & 87 & 189.45 \\
Validity WBF & 213 & 104 & 201.69 \\
\bottomrule
\end{tabular}
\end{table}

The optimized reductions preserve the reference float64 summation and
two-pass variance before float32 feature storage. Compiler settings prevent
reassociation and contraction. A signed-negative-zero input uses the
reference reduction path to retain platform-specific ties. Complete
numerical conventions and build settings accompany the code.

There are 432 compiled synthetic checks over empty through 1000-candidate
inputs, including sparse, dense, duplicate, tiny/degenerate, threshold,
signed-zero and reduction-boundary cases. Features and diagnostic IoUs
match exactly on the tested platform. The 64-image real-packet preflight
checks both meta heads at the feature, raw probability, scalar score,
stable rank, threshold-mask and final-output levels in each job.
Each formal request also matches the frozen ordered output. These
checks cover audited inputs and versions, not every compiler, operating
system or possible floating-point input.

Two independently initialized jobs reverse policy order over the same
192 image--repeat blocks. Each contains 1344 formal forwards across
seven policies; initialization, smoke, preflight and warmup bring the
overall forward count to 2888. Repeated measurements are averaged within
image before means/quantiles. They do not add independent images.
Source-specific means and components are below. Components include
nested clocks and should not be summed to reconstruct total time.

On this exact workload, Full100's quality is 925 TP/308 UP, raw
Hard's is 897/348, Meta138's is 922/311 and Meta141's is 924/309.
No subset AP is computed. The much larger CH/MOT cohort quality is
not attached to a 64-image latency number as a single population point.
Own-process Torch allocator peaks include detector residency; process
RSS and lifetime high-water marks cannot identify a clean marginal
memory peak for each policy.

\subsection{Earlier sessions and operator sensitivity}
Earlier measurements answer different questions and retain their own
images. In the same-Hard Timing-128 pair, raw takes 83.0146 ms and
Full100 87.0392 ms, with actual outputs 1291 TP/483 UP and 1315/515.
The additional 4.0246 ms includes capture and readout dependencies.
For the separately fitted RetinaNet, matched raw/quality WBF takes
182.51/195.07 ms, ratio 1.0688 (descriptive 95\% interval
[1.0610,1.0763]). Even identical WBF parameters permit scores to alter
clustering work and geometry. Neither session replaces the final
optimized-meta comparison or uses another cohort's best quality.

\section{Implementation Details}
Sections S1 and S2 specify candidate construction, feature conventions,
fitting, scalar mapping and output selection. Section S3 gives the data
roles and Appendixary results; Section S4 specifies the complete-time
boundary and execution platform.





\end{document}